\pdfoutput=1

\documentclass[11pt]{article}

\usepackage[final]{acl}

\usepackage{times}
\usepackage{latexsym}

\usepackage[T1]{fontenc}

\usepackage[utf8]{inputenc}

\usepackage{microtype}

\usepackage{inconsolata}

\usepackage{graphicx}

\title{Gender-Neutral Large Language Models for Medical Applications: Reducing Bias in PubMed Abstracts}

\author{Elizabeth Schaefer \\
  Yale University\\ Department of 
   Computer Science \\ New Haven, CT\\
  \texttt{Elizabeth.Schaefer@yale.edu} \\\And
  Kirk Roberts \\
  UTHealth Houston\\
  McWilliams School of 
  Biomedical \\ Informatics, Houston, TX\\
  \texttt{Kirk.Roberts@uth.tmc.edu} \\}

\begin{document}
\maketitle
\begin{abstract}
This paper presents a pipeline for mitigating gender bias in large language models (LLMs) used in medical literature by neutralizing gendered occupational pronouns. A set of 379,000 PubMed abstracts from 1965-1980 was processed to identify and modify pronouns tied to professions. We developed a BERT-based model, ``Modern Occupational Bias Elimination with Refined Training,'' or ``MOBERT,'' trained on these neutralized abstracts, and compared it with ``1965BERT,'' trained on the original dataset. MOBERT achieved a 70\% inclusive replacement rate, while 1965BERT reached only 4\%. A further analysis of MOBERT revealed that pronoun replacement accuracy correlated with the frequency of occupational terms in the training data. We propose expanding the dataset and refining the pipeline to improve performance and ensure more equitable language modeling in medical applications.
\end{abstract}

\section{Introduction}

\subsection{Background}

Large language models (LLMs) are now widely used for a range of applications, from creating customer service chatbots to advertising that targets specific clients to predicting financial outcomes from potential economic indicators. LLMs have also increased in presence in the medical sector, ranging from accessible diagnostics to comprehensive literature retrieval, where they hold the promise of leading to a more informed level of care. Given the critical nature of these uses, it is essential to ensure that such LLMs remain free from biases that could potentially impact patient treatment and outcomes.

Despite their potential, though, many LLMs have been shown to contain and perpetuate biases \cite{Kotek2023, Liu2022, Abid2021, Prakash2023, Bai2024, Zack2024, Bedi2024, Degelin2024}. The presence of these biases in LLMs is especially concerning in medical applications, where it can lead to incorrect diagnoses, inappropriate treatment recommendations, and ultimately, unequal healthcare. For example, an LLM fine-tuned on a dataset like PubMed might provide biased diagnostic suggestions if the underlying data contain gendered stereotypes. Gender biases and their effects have already been highlighted in a range of medical practice cases, for topics that include generalized surgical procedures, psychiatry, kidney transplantation, and intensive care treatment, among many other areas \cite{Ruiz1997, vanDaal2020, Lim2021, Merdji2023, Omar2024}. Our research focuses specifically on occupational bias in conjunction with gendered pronouns, highlighting the underrepresentation and exclusion of women from traditionally male-dominated professions, a critical area given the concomitant distortion that can result from that in patient care decisions. Numerous instances of this bias are evident in the PubMed dataset, as illustrated in Figure 1. Instead of perpetuating these inequalities, a properly and rigorously trained LLM can mitigate and avoid such dangerous generalizations. 

\begin{figure*}[h]
    \centering
    \includegraphics[width=0.6\linewidth]{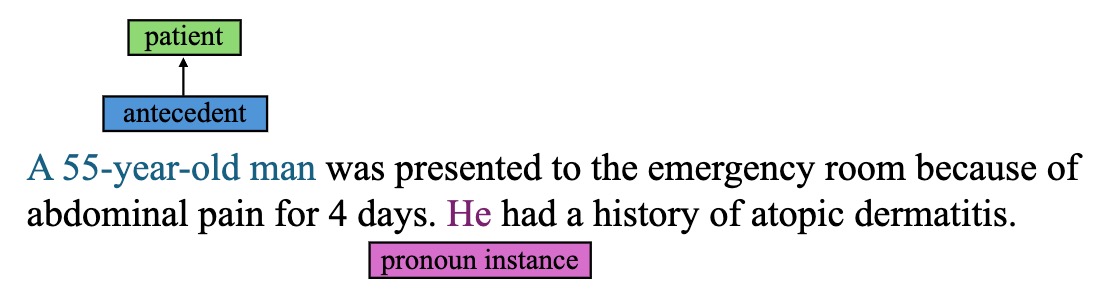}
    \includegraphics[width=0.7\linewidth]{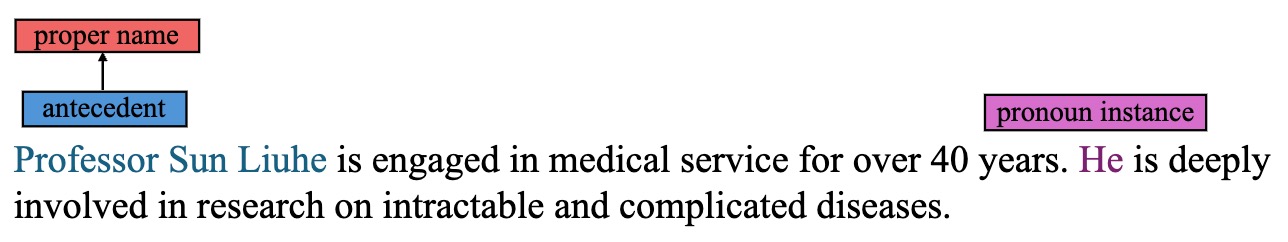}
    \includegraphics[width=0.6\linewidth]{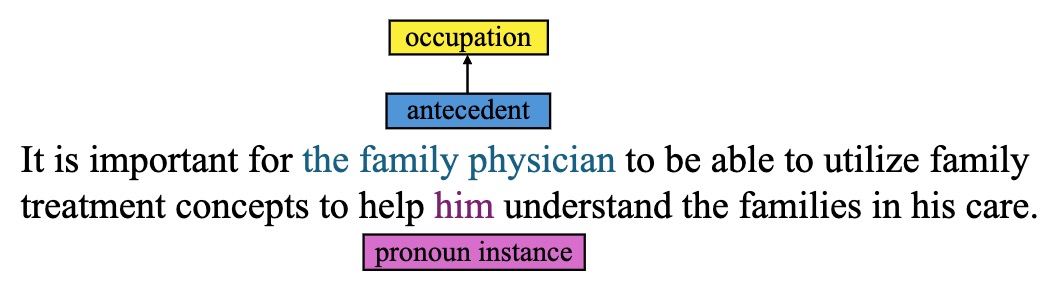}
    \includegraphics[width=0.7\linewidth]{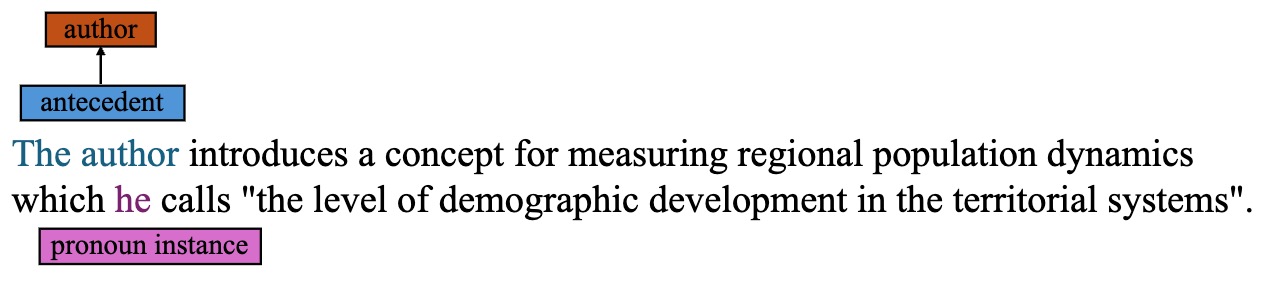}
    \caption{Example annotations from PubMed abstract text.}
    \label{fig:fig1}
    \vspace{-0.1in}
\end{figure*}

In this paper, we propose a novel approach to mitigate gender bias in LLMs used in medical contexts. To ensure that only relevant pronouns are neutralized without affecting critical medical details, our pipeline specifically targets pronouns that refer solely to occupations. This process preserves medically significant context, such as patient-specific information, while eliminating biased language tied to occupational stereotypes.

Our method focuses on addressing biases in the training data before the fine-tuning stage of LLM development. By constructing and validating a robust query pipeline that identifies and neutralizes binarily-gendered pronouns linked to occupational terms in medical literature, we aim to create more reliable and fair models. This aligns with concerns raised by \newcite{bender2021dangers}, who emphasize that training on biased corpora can amplify existing societal stereotypes in unintended ways. By modifying the dataset at the source rather than post-processing model outputs, we directly address these concerns and create a more stable foundation for fairness in medical NLP. Our pipeline includes several key components: a lexicon for identifying gender-specific pronouns, a pronoun resolution query, and a classification query to identify occupation-specific subjects. Both queries were conducted using Meta's Llama-3.1-405b model \cite{meta2024llama3.1}, which we elected to use because it was the most advanced Llama model available at the time, and offered improved reasoning and accuracy for the task compared to previous models. The effectiveness of this approach is demonstrated through the development and evaluation of a BERT-based model, ``Modern Occupational Bias Elimination with Refined Training,'' or ``MOBERT,'' trained on gender-neutralized abstracts from PubMed.

\subsection{Related Work}
The issue of bias in word embeddings and LLMs has been widely studied \cite{Pessach2022}, with researchers highlighting how models trained on human-generated corpora often reflect and amplify societal stereotypes \cite{Dev2023, Ungless2022}, as well as proposing both technological and social solutions. \newcite{Bolukbasi2016} first demonstrated that word embeddings could capture and propagate gender biases, showing that terms like ``programmer'' were more closely associated with men than women. Subsequent research provided further examples of such biases and explored their widespread implications in the field \cite{Ray2023, Bommasani2021, Mehrabi2021, An2024, Pervez2024}. To counter this effect, \newcite{Bolukbasi2016} proposed a post-processing technique to debias word embeddings by projecting gender-neutral words into a subspace orthogonal to a gender direction. However, their method had limitations, such as requiring a classifier to identify gender-neutral words, which could introduce errors and propagate bias if the classifier itself was flawed. 

\newcite{Zhao2018} subsequently introduced GN-GloVe, a method that embeds gender information into specific dimensions of word vectors while neutralizing others. This approach improved interpretability and allowed for more effective debiasing by focusing on protected attributes like gender. However, while GN-GloVe effectively reduced direct gender bias, it left room for improvement in terms of indirect bias and applicability to contextualized word embeddings. As \newcite{blodgett-etal-2020-language} highlight, many prior bias-mitigation efforts rely on lexicon-based heuristics or simple vector-space transformations, which may fail to generalize to real-world applications. Our work moves beyond this by integrating pronoun resolution and context-aware classification before model fine-tuning.

More recently, the focus has shifted to contextualized word embeddings, such as those used in LLMs like BERT and GPT. \newcite{Basta2019} and \newcite{Zhao2019} explored gender bias in these models, finding that while contextualized embeddings reduce some biases present in static embeddings, they still retain significant levels of bias, especially in how they handle occupations and pronouns in context. In line with these concerns, recent work has shifted attention toward non-binary and gender-neutral pronouns, as more individuals identify outside the binary gender framework. Although much of the previous research focused on binary gender categories, studies such as \newcite{Hossain2023} have revealed that large language models struggle significantly with gender-neutral and neo-pronouns, like ``they'' or ``xe.'' This highlights the broader issue of representation in training corpora, where non-binary pronouns are often underrepresented, exacerbating the model’s difficulty in handling inclusive language effectively.

Beyond post-processing and embedding-based debiasing methods, recent research has explored direct modifications to training data as a strategy for mitigating gender bias in biomedical NLP.
\newcite{Agmon2024} introduced TeDi-BERT, a model that applies temporal distribution matching to adjust how gender is represented in historical clinical trial abstracts, ensuring that language models trained on older data align more closely with contemporary gender distributions. Their approach highlights the importance of modifying training corpora before model training to prevent biased language from propagating in downstream applications.

Our work builds on these foundations but diverges in its focus on occupational bias in medical literature. Rather than aligning embeddings across different time periods, we develop a pipeline that systematically identifies and neutralizes gendered occupational pronouns before model fine-tuning. Through this methodology, we aim to create LLMs that are not only less biased but also more effective in delivering equitable healthcare outcomes. Unlike previous efforts that focused on post-processing or debiasing at the embedding level, our approach integrates bias mitigation into the model training process, addressing both direct and indirect biases more comprehensively.

\begin{table*}[t]
\caption{Descriptive statistics of the annotated corpora.}
\vspace{-0.1in}
\label{tab:table1}
\footnotesize
%
\begin{tabular}{|l|l|l|}
\hline
  & \textbf{Dataset A: Pronoun-only corpus} & \textbf{Dataset B: Pronoun- and Lexicon-Derived Corpus}\\
\hline
Total Number of Abstracts & 250 & 500 \\
\hline
Patient/Trial Participant &	28 & 323 \\
\hline
Named Individual & 62 & 115 \\
\hline
Occupation & 97 & 24 \\
\hline
Author of the Abstract & 56 & 19 \\
\hline
Animal & 0 & 7 \\ 
\hline
Other & 7 & 12 \\
\hline
\end{tabular}%
\end{table*}

\section{Methods}
\subsection{Data}
Our data are taken from the MEDLINE 2019 baseline set of PubMed abstracts from 1965 through mid-2018, totaling approximately 29 million abstracts. We utilized two lexicons to locate relevant abstracts for our study. The first lexicon searched for binarily-gendered pronouns, such as ``him,'' ``her,'' and ``himself,'' ensuring that only complete words were identified. This process reduced the initial set of 29 million abstracts to 687,000 relevant abstract instances. A second lexicon, designed to identify occupational terms, was applied only for testing purposes, allowing us to evaluate classification performance across a broader range of occupation-related pronouns. This second lexicon was not used in the case study dataset and did not affect the pronoun-neutralization process. For clarity, we designate the “Pronoun-only corpus” as Dataset A and the “Pronoun- and Lexicon-Derived Corpus” as Dataset B. These names will be used throughout the paper when referring to the annotated datasets, as in Table~\ref{tab:table1}.

Each instance in our corpus represents the character offset of each pronoun found within an abstract. This means that if an abstract contains three gendered pronouns, it will appear in our baseline corpus three times, once for each pronoun occurrence. This approach is crucial for determining the specific pronoun resolution in each instance, as different resolutions may occur within a single abstract.

\begin{table*}[t]
\caption{Categorization rules for classifying an antecedent within the context of an abstract.}
\vspace{-0.1in}
\label{tab:table2}
\footnotesize
\begin{tabular}{|p{1.3in}|p{4.6in}|}
\hline
\textbf{Antecedent category} & \textbf{Category definition} \\
\hline
Patient/Trial Participant & Individuals directly receiving medical care, those with a medical condition, or who are injured. This also applies to any trial participant, defined as someone who volunteers or is examined in a study, regardless of whether they are an occupational subject or not. The label ``patient'' takes precedence over any other classification when applicable. \\
\hline
Named Individual        & Individuals referred to by a proper personal name, which includes capitalized names or redacted names. \\
\hline
Occupation           & Individuals, real or abstract, identified by their profession or job, where they are employed and paid for their work.                                                                   \\
\hline
Author of the Abstract    & An author of the paper.               \\
\hline
Animal                    & Any non-human creature        \\
\hline
Other                     & Any instance that does not fit into the previous categories. \\                    
\hline
\end{tabular}%
\end{table*}

\subsection{Annotation Process}
After constructing Datasets A and B, we proceeded with a two-step annotation process involving pronoun resolution and antecedent classification. This annotation process (utilizing LabelStudio \cite{LabelStudio}) involved first identifying the noun phrase to which the pronoun referred (defined as “pronoun resolution”) and then classifying that antecedent within the context of the abstract according to the established classification rules. Those rules were set as seen in Table~\ref{tab:table2}. In this study, we intentionally avoided neutralizing pronouns when referring to patients or trial participants, as well as in contexts where biological sex is medically relevant. Certain conditions, such as prostate or ovarian cancer, are inherently gendered, and so de-gendering such references could hamper a model’s medical reasoning. Consequently, whenever a pronoun refers to a particular patient or group of patients in the abstract, that pronoun was left unchanged. Two annotators (the first author and an intern in the lab) separately labeled each given corpus, then calculated Cohen’s Kappa, a measure quantifying the level of agreement before reconciliation. Example annotations from the corpus are shown in Figure~\ref{fig:fig1}. Descriptive statistics of the annotated corpora from Dataset A and Dataset B are provided in Table~\ref{tab:table1}.

\subsection{Pronoun Resolution Query}
As the first step in our automated pipeline, we used a Llama-3.1-405b query to determine the subject associated with each pronoun, a process known as pronoun resolution. This step involved determining the noun or noun phrase to which a given pronoun referred within the abstract’s context, ensuring that additional descriptive clauses were excluded. The full structured prompt used in this query is detailed in Table~\ref{tab:table4}. To evaluate this prompt-based model, a randomly selected corpus of 500 pronoun instances was chosen from the relevant abstracts and each pronoun’s respective antecedent was located and double-annotated with a Cohen’s Kappa of 0.9000. Selected examples and the overall makeup of this corpus can be seen in both Figure~\ref{fig:fig1} and Table~\ref{tab:table2}. 

\subsection{Lexicon Validation}
Following pronoun resolution, it proved useful to define a mechanism to distinguish occupational antecedents from other noun phrases. While the LLM was highly effective in identifying pronoun antecedents, we found that only a small percentage of gendered pronouns were actually attributable to professions, the core focus of our task. This data imbalance made it difficult to obtain a sufficient sample of occupation-related pronoun instances for evaluation. To address this, we developed a lexicon specifically designed to increase the frequency of identified occupational antecedents. The lexicon was initially derived from the synset relations of ``professional'' in WordNet, incorporating common occupational terms and case-sensitive acronyms (e.g., ``rn'' vs. ``RN''). To validate the lexicon’s efficacy, it was applied to the 500 resolved antecedents, filtering for occupational terms. 

\subsection{Classification Query}
Using the validated lexicon described above, the results of our antecedent query can be successfully filtered. In this application, 250 pronoun instances were extracted from the data (primarily from Dataset B with the applied lexicon), along with their corresponding antecedents, that included occupational terms. These 250 instances and the text of the abstract in which they appeared must also be examined and tested for accuracy. Those instances were double-annotated and reconciled with a Cohen’s Kappa of 0.9470. After annotator reconciliation, we used the Llama-3.1-405b model to classify each antecedent according to the same labeling rules, enabling comparison between human and model performance.

\begin{table*}[t]
\caption{The pipeline performance for the classification query.}
\vspace{-0.1in}
\label{tab:table3}
\begin{center}
\footnotesize
\begin{tabular}{|l|l|l|l|l|}
\hline
\textbf{Annotation}    & \textbf{\begin{tabular}[c]{@{}l@{}}Frequency \\ (out of 250)\end{tabular}} & \textbf{Precision} & \textbf{Recall} & \textbf{F1} \\ \hline
Occupation             & 97                                                                         & 0.9895             & 0.9691          & 0.9792      \\ \hline
Named Individual       & 62                                                                         & 0.9492             & 0.9032          & 0.9252      \\ \hline
Author of the Abstract & 56                                                                         & 1                  & 0.9107          & 0.9533      \\ \hline
Patient                & 28                                                                         & 0.7027             & 0.9286          & 0.8         \\ \hline
Other                  & 7                                                                          & 0.75               & 0.8571          & 0.8         \\ \hline
Macro Weighted Avg. &                                                                            & 0.943              & 0.932           & 0.9349      \\ \hline
\end{tabular}
\end{center}
\end{table*}

\begin{table*}[t]
\caption{The Llama-3.1-405b prompts for the pronoun resolution and classification queries.}
\vspace{-0.1in}
\label{tab:table4}
\begin{center}
\footnotesize
\begin{tabular}{|p{1.2in}|p{2in}|p{2.5in}|}
\hline
  & \textbf{System Content}                                                                                                                                    & \textbf{User Content}                                                                                                                                                                                                                                                                 \\ \hline
  \textbf{Pronoun Resolution Query}
                                                 & You are a helpful assistant with identifying the direct antecedent of a pronoun. Here is your antecedent\_background knowledge: \{background\}'*. & Identify the direct antecedent of the pronoun marked with {[}START{]} and {[}END{]} in the following abstract: \{highlighted\_abstract\}. Only answer with the antecedent.                                                                                                   \\ \hline
\textbf{Classification Query (where {antecedent} is the output of Antecedent Query)} & You are a helpful assistant following these classification rules \{rules\}.**                                                                     & In the following abstract, classify which label the noun ``\{antecedent\}'' in the context of the abstract \{highlighted\_abstract\} is referring to:   ``patient,'' ``occupation,'' ``named individual,''   ``author,'' ``animal,'' or ``other.'' Only output the   label, no other text. \\ \hline
\end{tabular}
\end{center}
\scriptsize
*This background information consists of antecedent grammatical rules established by Fordham \cite{Fordham2024}. \\
{**}These classification rules consist of the same rules shown in the Annotation Process section.
\vspace{-0.8cc}
\end{table*}

\subsection{Pronoun Neutralization Process }
After identifying gendered pronouns linked to occupational terms, the next step was neutralization. NLTK was used for tokenization and part-of-speech tagging.
A pronoun-mapping dictionary was developed to replace gendered pronouns with their gender-neutral counterparts, such as `they/them/theirs.'
This dictionary accounted for compound pronouns (e.g., \textit{he or she} $\rightarrow$ \textit{they}) and handled replacements while preserving sentence structure.
Pronouns flagged for neutralization were modified only when they referred to occupational antecedents, ensuring no changes were made to pronouns referring to patients or trial participants.
This distinction was critical for maintaining medically relevant context in abstracts where sex-specific conditions (e.g., prostate cancer) were discussed.
Examples of pronoun replacements and contexts that were preserved are presented in Table~\ref{tab:table5}.
The following section presents the results obtained from applying the three-stage pipeline on our annotated datasets, showcasing the effectiveness of our approach in neutralizing gendered pronouns.

\begin{table*}[t]
\caption{Examples of phrases that would/would not be identified for replacement, and the resulting modifications.}
\vspace{-0.1in}
\label{tab:table5}
\begin{center}
\footnotesize
    \begin{tabular}{|p{2.2in}|p{0.9in}|p{0.8in}|p{1.9in}|}
\hline
\textbf{Example Sentence}                  & \textbf{Antecedent} & \textbf{Label} & \textbf{Modification} \\ \hline
Some compromise must be reached between the unwillingness of the surgeon to spend most of \textbf{his} time performing abortions and the freedom for women to have them. 

[PMID: 5598532, 10/25/1968]  & the surgeon    & Occupation & Some compromise must be reached between the unwillingness of the surgeon to spend most of \textbf{their} time performing abortions and the freedom for women to have them. \\ \hline
 Before any physician attempts to treat telangiectasia by this method, \textbf{he or she} should observe its performance by an experienced operator. 
 
 [PMID: 834688, 3/15/1977] & any physician & Occupation & Before any physician attempts to treat telangiectasia by this method, \textbf{they} should observe its performance by an experienced operator. \\ \hline
 Four lectures given by Dr. Mora and \textbf{his} staff focus on the betterment of the quality of life through improved nutrition. 
 
 [PMID: 12261512, 6/10/1980] & Dr. Mora           & Proper name    & No modification.      \\ \hline
\end{tabular}
\end{center}
\vspace{-0.1in}
\end{table*}

\section{Results}
\subsection{Pipeline Performance Metrics}
To evaluate the performance of our pipeline, we analyzed two separate annotated datasets. Dataset A (Pronoun-only corpus) was used to assess the pronoun resolution component, while Dataset B (Pronoun- and Lexicon-Derived corpus) was utilized for the lexicon validation and classification queries. 
First, using Dataset A, we applied our pronoun resolution query on the non-annotated corpus. The Llama-3.1-405b query was run and the resulting pronoun instances and their corresponding antecedent outputs were cross-referenced with the ground-truth annotations. This comparison yielded an accuracy of 0.9881 on the initial 500 abstracts, demonstrating that our pronoun resolution method reliably identifies antecedents. Next, with Dataset B, we validated our lexicon for identifying occupational terms by applying it to the 500 antecedents obtained from the pronoun resolution query. The filtered results were then compared with the `occupation' labels in the ground-truth annotations, achieving a perfect recall score of 1.0000. This confirms that our lexicon effectively identifies occupational antecedents for the classification task.
Finally, still using Dataset B, we assessed the performance of our Llama-3.1-405b classification query by calculating precision, recall, and the F1 score between the generated labels and the ground-truth labels. The numerical outcomes of this process are presented in Table~\ref{tab:table3}, and an example of an antecedent versus classification query is provided in Table~\ref{tab:table4}. These results confirm the high performance of our classification approach in accurately distinguishing occupational pronoun instances from other categories. Together, these performance metrics validate the robustness of our pipeline, linking each methodological step to successful outcomes in resolving, filtering, and classifying gendered pronoun instances.

\subsection{Pronoun Neutralization Case Study}
We tested the effect of our pipeline on a corpus of the 379,000 PubMed abstracts from 1965-1980, hypothesizing that these texts would show a greater prevalence of singular gendered pronouns, based on a qualitative examination of a random sample set of the abstracts. After processing this corpus, pronouns linked to occupational antecedents were neutralized in 1,400 abstracts.

To determine the success of this replacement, we trained two separate base uncased BERT models \cite{Devlin2019}. The first model, named 1965BERT, was trained on the original, unmodified dataset of the 379,000 PubMed abstracts from 1965-1980. The second model, denoted ``Modern Occupational Bias Elimination with Refined Training,'' or MOBERT, was trained on a similar dataset of the 379,000 abstracts, but with 1,400 abstracts identified and modified with the newly introduced gender-neutral tokens. Additionally, these 1,400 abstracts were analyzed to identify the most frequently-occurring occupational terms in relevant antecedents. The top five terms identified were ``physician,'' ``surgeon,'' ``doctor,'' ``practitioner,'' and ``nurse.'' Both models were trained for three epochs with a batch size of four per device, using a mixed precision (fp16) configuration across multiple graphics processing units. Training logs were saved at regular intervals, with models checkpointed every 10,000 steps.

To further assess the models, we conducted a masked language modeling test using 50 sentences from our initial annotated corpus of 500 abstracts, ensuring that each randomly selected sentence contained gendered pronouns from post-1980 texts. Importantly, the models were not trained on the data used in these tasks, ensuring an independent evaluation of their performance. The testing corpus was assembled by selecting ten sentences for each of the five most frequent occupational terms identified, resulting in 50 sentences. In each sentence, a [MASK] token was inserted in place of a pronoun, and the model was tasked with predicting the correct pronoun when given respective options of `he/him/his,' `she/her/hers,' and `they/them/theirs.'

\begin{table*}[t]
\caption{Examples of the masking test and the corresponding outcomes.}
\vspace{-0.1in}
\label{tab:table6}
\begin{center}
\footnotesize
    \begin{tabular}{|p{1.8in}|p{0.9in}|p{1.1in}|p{0.8in}|p{0.8in}|}
\hline
\textbf{Example Sentence}                  & \textbf{BERT-Base} & \textbf{PubMedBERT} & \textbf{1965BERT} & \textbf{MOBERT} \\ \hline

Although a doctor may not be continually aware of it, [MASK] medical activity is firmly rooted in the moral principles of the medical profession.

PMID: 7470698, 5/21/1981
& his    & his & his & their \\ \hline

Many different portable computers are currently available and it is now possible for the physician to carry a mobile computer with [MASK] all the time.

PMID: 12835877, 8/29/2003
& them & him & him & them \\ \hline

\end{tabular}
\end{center}
\end{table*}

\subsection{Outcomes}
We compared the results of this masking test between BERT-Base (the untrained model), PubMedBERT, 1965BERT, and MOBERT (all three of which are trained upon BERT-Base with their respective training data) \cite{microsoft}. Examples of this masking test and the corresponding outcomes are in Table~\ref{tab:table6}, with overall results shown in Table~\ref{tab:table7}. Percentages indicate the proportion of sentences in which gender-inclusive pronouns (`they/them/theirs') replaced gendered pronouns. For example, if BERT-Base replaces 40\% of masked pronoun instances with a gender inclusive pronoun, 1965BERT replaces 4\% of those same instances with a gender inclusive pronoun. The MOBERT results were further analyzed to determine a relationship between the frequency of the occupational term in the training data and the accuracy of replacement, as shown in Table~\ref{tab:table8}.

\section{Discussion}
\subsection{Principal Results}
The application of our gender-neutralization pipeline to the 1965-1980 PubMed abstracts has demonstrated its potential to significantly reduce occupational gender bias in large language models. By introducing gender-neutral pronouns reconciled with occupational terms in 1,400 abstracts, we successfully trained a model, MOBERT, that demonstrated a 70\% success rate in predicting inclusive pronouns in a masked language modeling task. This result far exceeds the 4\% success rate of 1965BERT, a model trained on unmodified texts from the same period, and highlights the importance of correcting biased data at the training stage. MOBERT's performance also surpassed that of both the base model, BERT-Base, which exhibited a 40\% success rate, and PubMedBERT, a model trained on the complete PubMed dataset without gender-neutralization, which achieved only a 20\% inclusive successive rate. These comparisons underscore the critical role of targeted intervention in mitigating bias in language models.

\begin{table*}[]
\caption{Overall results for the inclusive replacement rates by model.}
\vspace{-0.1in}
\label{tab:table7}
\begin{center}
\footnotesize
\begin{tabular}{|l|l|}
\hline
\textbf{Model}                                & \textbf{Inclusive Replacement Rate (\%)} \\ \hline
BERT-Base (Comparison Baseline for All Cases) & 40                                     \\ \hline
PubMedBERT                                  & 20                                     \\ \hline
1965BERT                                      & 4                                      \\ \hline
MOBERT                                    & 70                                    \\ \hline
\end{tabular}
\end{center}
\end{table*}

\subsection{Comparison with Prior Work}
Research on the recruitment and retention of women in male-dominated occupations highlights how deeply embedded gendered language can reinforce exclusionary workplace cultures \cite{Germain2012}. Prior studies such as \newcite{de-arteaga2019bias} have shown that occupational gender bias in machine learning software can directly affect hiring and professional representation. Research has shown that the assumption of male dominance in professional fields can discourage the participation of women in those fields, leading to self-reinforcing cycles of underrepresentation \cite{Wu2022}. For instance, studies have demonstrated that increasing gender diversity in male-dominated academic settings leads to improved career outcomes for female students, suggesting that removing implicit assumptions – such as assuming doctors are male – could encourage more diverse participation in medicine, technology, engineering, and math \cite{Germain2012}. If gendered language in professional texts perpetuates the underrepresentation of women in these fields, systematically neutralizing such biases could contribute to breaking this cycle.

\subsection{Future Improvements}
Future studies could involve integrating MOBERT into clinical NLP applications – such as diagnostic models and medical literature retrieval systems – to assess whether gender neutralization leads to improved healthcare equity. \newcite{sun-etal-2019-mitigating} highlight that debiasing techniques should be evaluated not only through linguistic performance but also through real-world impact within medicine. Conducting user studies with medical professionals would be useful in assessing how gender-neutral models influence literature search relevance and clinical decision-making.

\begin{table*}[t]
\caption{Relationship between the occupational term frequency in the training data and the replacement accuracy.}
\vspace{-0.1in}
\label{tab:table8}
\begin{center}
\footnotesize
    \begin{tabular}{|l|l|l|}
\hline
\textbf{Occupational Term} & \textbf{Frequency} & \textbf{Percentage (\%)} \\ \hline
Physician                  & 298                & 100               \\ \hline
Surgeon                    & 135                & 100              \\ \hline
Doctor                     & 89                 & 70                \\ \hline
Practitioner               & 68                 & 60               \\ \hline
Nurse                      & 64                 & 30               \\ \hline
\end{tabular}
\end{center}
\vspace{-1cc}
\end{table*}

\section{Conclusions}
This work demonstrates the effectiveness of a gender neutralization pipeline in reducing occupational gender bias in large language models trained on medical literature. By processing 379,000 PubMed abstracts from 1965-1980 and targeting gender-specific pronouns linked to professions, we improved MOBERT’s success rate to 70\% in predicting gender-neutral pronouns, compared to 4\% for 1965BERT. This improvement highlights the importance of addressing bias during training. While promising, the study also reveals opportunities for improvement, such as expanding the dataset and integrating the pipeline into further applications. These findings underscore the potential for creating more equitable and unbiased models in medical and other sensitive domains.

\section*{Limitations}

Despite the overall success of MOBERT, our analysis did reveal some limitations. As seen in Table 8, the frequency of occupational terms in the 379,000 modified abstracts used for pre-training correlated strongly with the accuracy of pronoun replacement. For instance, terms like “physician” and “surgeon,” which appeared more frequently in the training data, saw a 100\% accuracy in neutral pronoun predictions, while terms like ``nurse'' had a much lower replacement rate of 30\%. Although we did not use supervised fine-tuning, MOBERT’s exposure to gender-neutralized occupational terms during pre-training likely contributed to its improved performance on those terms in masked language modeling tasks. Since masked language modeling relies on contextual co-occurrence rather than explicit supervision, MOBERT likely developed stronger associations between certain occupations and gender-neutral pronouns due to their repeated exposure in the training data. This suggests that expanding a pre-training dataset to include a broader range of occupations and more balanced representation of male- and female-dominated roles could further improve the model’s performance.

Another potential limitation arises from the process of language alteration itself. Although we carefully designed our pipeline to neutralize pronouns only in contexts where the occupational term was the antecedent, there remains a risk that some instances of gendered language with medically significant context may have been inadvertently modified. While we found no evidence of such errors in our testing, further refinements to the pipeline could incorporate more sophisticated contextual analysis to ensure the protection of patient-specific or trial-related information. Additionally, large-scale gender neutralization poses the challenge of maintaining critical semantic distinctions. While replacing a gendered pronoun with `they/them/theirs' often preserves meaning, certain contexts – such as historical citations or patient narratives – could lead to unintended distortions. Our lexicon filtering approach helps mitigate this by restricting modifications to occupational contexts; however, broader applications must carefully handle edge cases where neutralization may introduce ambiguity or alter medically relevant details. For example, in a PubMed abstract (PMID: 25549443) there is a sentence discussing gender dynamics in surgery: \textit{``Suggestions include a change [in] the relationship between a female surgeon and her partner, a supplement of surgeons so that hospitals could change the traditional system of surgery.''} Indiscriminate neutralization of \textit{``her''} to \textit{``their''} could obscure the focus on challenges specific to female surgeons, weakening the text’s emphasis on gendered professional and personal expectations. Addressing these concerns will require further refinement, including human evaluation studies, dependency parsing for syntactic precision, and additional lexicon constraints to safeguard against unintended language shifts.

\section*{Acknowledgements}
Elizabeth Schaefer was supported by the Cancer Prevention and Research Institute of Texas (CPRIT RP210045) as part of the Biomedical Informatics, Genomics and Translational Cancer Research Training Program. Kirk Roberts was supported by the National Institutes of Health (R01LM014508, R01LM011934). The authors acknowledge Dhruv Kumar of the University of Texas for serving as a second annotator in the data labeling process, contributing to the double-annotation and reconciliation of the corpus.

\bibliography{amiabib}

\end{document}